\documentclass[a4paper,twoside]{article}

\usepackage{epsfig}
\usepackage{subfigure}
\usepackage{calc}
\usepackage{amssymb}
\usepackage{amstext}
\usepackage{amsmath}
\usepackage{amsthm}
\usepackage{multicol}
\usepackage{pslatex}
\usepackage{apalike}

\usepackage{graphicx}
\usepackage{tikz}
\usetikzlibrary{calc}
\usetikzlibrary{chains}
\usetikzlibrary{shapes}
\usetikzlibrary{decorations.pathreplacing}
\usepackage{pgfplots}
\pgfplotsset{compat=newest}

\DeclareMathOperator{\argmax}{argmax}

\usepackage{amsmath,amssymb,amsfonts}
\usepackage{algorithmic}
\usepackage{algorithm}
\usepackage{textcomp}
\usepackage{xcolor}
\usepackage{url}
\usepackage{hyperref}
\usepackage{SCITEPRESS}     
\newcommand{\figref}[1]{Fig.~\ref{#1}}
\newcommand{\Tiblref}[1]{Table~\ref{#1}}
\newcommand{\secref}[1]{Section~\ref{#1}}

\begin{document}

\title{Adversarial Alignment of Class Prediction Uncertainties for Domain Adaptation}
\author{\authorname{Jeroen Manders\sup{1,2}, Twan van Laarhoven\sup{1,3} and Elena Marchiori\sup{1}}
\affiliation{\sup{1}Institute for Computing and Information Science, Radboud University, Nijmegen, The Netherlands}
\affiliation{\sup{2}TNO, The Netherlands}
\affiliation{\sup{3}Faculty of Management, Science and Technology, Open University, Heerlen, The Netherlands}
\email{jeroenmanders89@gmail.com, mail@twanvl.nl, elenam@cs.ru.nl}
}


\abstract{We consider unsupervised domain adaptation: given labelled examples from a source domain and unlabelled examples from a related target domain, the goal is to infer the labels of target examples. Under the assumption that features from pre-trained deep neural networks are transferable across related domains, domain adaptation reduces to aligning source and target domain at class prediction uncertainty level.  
We tackle this problem by introducing a method based on adversarial learning which forces the label uncertainty predictions on the target domain to be indistinguishable from those on the source domain.  
Pre-trained deep neural networks are used  to generate deep features having high transferability across related domains. 
We perform an extensive experimental analysis of the proposed method over a wide set of publicly available pre-trained deep neural networks. 
Results of  our experiments on domain adaptation tasks for image classification show that class prediction uncertainty alignment with features extracted from pre-trained deep neural networks provides an efficient, robust and effective method for domain adaptation. }

\onecolumn \maketitle \normalsize \vfill

\section{INTRODUCTION}\label{introduction}

In unsupervised domain adaptation,  labelled examples from a source domain and unlabelled examples from a related target domain are given. The goal is to infer the labels of target examples. 
A straightforward approach for tackling this problem is to label target examples by just applying a deep neural network pre-trained on data from a related domain. This approach has been shown to work rather well in practice. The reason is that  deep networks learn feature representations which reduce domain discrepancy, although they do not fully eliminate it \cite{yosinski2014transferable}. 
   If we assume that features from pre-trained deep neural networks indeed provide a good representation for both source and target data, then in order to perform domain adaptation one needs to align only source and target label predictions in such representation, so only at class label level.   This paper investigates this novel setting.
We introduce a label alignment  method  to  force the uncertainty in the predicted labels on the target domain to be indistinguishable from that on the source domain. The method is based on the adversarial learning approach: it performs domain alignment at class label level  by learning a representation that is at the same time discriminative for the labelled source data yet incapable of discriminating the source and target at prediction uncertainty level. Specifically, the proposed method considers the class probabilities of a label classifier $f$ {\it as input of a domain discriminator} $g$. The label classifier $f$ is trained to minimize the standard supervised loss on the source domain while the domain discriminator $g$ is trained to distinguish the class probabilities that the label classifier $f$ outputs on the source domain from those on the target domain (see \figref{fig:architecture}).

A limitation of this method is that it works only under the assumption that source and target domains have the same class distribution. This is because the representation chosen to minimize the discrepancy between domains depends on the domains class distribution. If they are different, then the discrepancy between the domains in our representation will be large. 

To overcome this limitation, we introduce a tailored loss function to enforce the domains to have equal class distributions during our training procedure: we incorporate  class weights in our loss function, one for each instance. Class weights of source examples are fixed and those of the target examples are updated during optimization of our loss function.  

Interestingly, training with this loss function leads to an overestimation of the target domain predictions, which results in an increased loss while stabilizing accuracy. This approach favors robustness, because the domain discriminator punishes overconfidence on the source domain, the latter being a sign of overfitting.

We call the resulting domain adaptation method LAD (Label Alignment with Deep features).

LAD  uses deep features extracted from a pre-trained deep neural network, so no fine tuning of the feature extractor is needed. As such, LAD is more efficient than end-to-end deep learning adaptation methods.

 Joint efforts from the machine learning research community resulted in the public availability of different pre-trained deep neural network architectures for visual classification. These pre-trained models  provide a rich variety of feature extractors. 
Besides trying to align label predictions, the other contribution of this paper is an extensive analysis of the same method when changing the `feature extractor' part by exploring a wide set of existing pre-trained deep architectures. This choice allows us to understand the advantage of the approach when dealing with different features.

An extensive experimental analysis shows that LAD achieves consistent improvement in accuracy  across different pre-trained networks.
Overall the method achieves state of the art results on the standard Office-31 and ImageCLEF-DA datasets, with a neat improvement over competing baselines on harder transfer tasks.

The main contributions of this paper can be summarized as follows: 1) a specific setting for the domain adaptation problem; 2) a tailored method  for performing domain adaptation in this setting; 3) extensive analysis of the proposed method when changing its `feature extraction' part.

\section{RELATED WORK}\label{relatedwork}
There is a vast literature on domain  adaptation (see for instance the recent surveys \cite{Weiss2016,csurka2017domain}).

Besides the optimization towards better source domain class predictions,  domain adaptation methods try to achieve domain invariance \cite{ben2010theory}.
To achieve domain invariance there are roughly two popular approaches: minimizing some measure of domain discrepancy and using an adversarial domain discriminator. Below we summarize several recent methods based on deep neural networks.

Deep-CORAL \cite{sun2016deep} aligns correlations of layer activations in deep neural networks.
Deep Transfer Network (DTN) \cite{zhang2015deep} employs a deep neural network to model and match both the domains marginal and conditional distributions. 
A popular measure used to minimize domain discrepancy is Maximum Mean Discrepancy (MMD) \cite{gretton2009covariate}.
This measure is used in several recent domain adaptation methods. For instance, 
DAN \cite{long2015learning} and RTN \cite{long2016unsupervised} are end-to-end deep adaptation methods which use the sum of multiple MMD by matching the feature distributions of multiple layers across domains.

In adversarial learning a generator tries to fool a discriminator so that it cannot distinguish between generated and real examples\cite{goodfellow2014generative}. 
Current work on adversarial domain adaptation tries to trick a domain discriminator so that it no longer can distinguish between features originating from either the source or target domain, which results in domain invariant features to be trained.

For instance, ReverseGrad \cite{ganin2015unsupervised,ganin2016domain} enforce the domains to be indistinguishable by reversing the gradients of the loss of the domain classifier.
 
Recently \cite{Tzeng_2017_CVPR} introduced a unifying framework for adversarial transfer learning, and proposed a new instance of this framework, called Adversarial Discriminative Domain Adaptation (ADDA), which combines discriminative modeling, untied weight sharing, and a generative adversarial network loss.

LAD's alignment at label level is based on adversarial learning. As such, it shares the theoretical motivation of adversarial learning method for domain adaptation, as explained e.g. in \cite{ganin2016domain}. 

Note that the idea of matching the classifier layer has been used in  end-to-end domain adaptation methods based on deep learning, for instance DAN and RTN, where both the feature layer and the classifier layer are aligned simultaneously using either MMD or adversarial domain discriminator. The main difference between LAD and these works is the underlying scenario: LAD works under the assumption that domain representations are already reasonably matched through the use of a pre-trained deep neural network for feature extraction. Therefore LAD performs alignment only at class label level, by matching label predictions.  While in LAD the domain discriminator takes as input (source and target) predictions of the label classifier,  in all previous adversarial methods for domain adaptation the domain discriminator takes as input (source and target) features.

\section{CLASS PREDICTION UNCERTAINTY ALIGNMENT}
\label{CPUA}


\renewcommand{\S}{\mathcal{S}}
\newcommand{\Ti}{\mathcal{T}}
\newcommand{\f}{{C}}
\newcommand{\g}{{D}}
\newcommand{\x}{\mathbf{x}}

We are given a set $\mathcal{S}$ of source images and their labels drawn from a source domain distribution $\mathbb{P}_S(\x,y)$,  and a set $\mathcal{T}$ of target images \emph{without} their labels, drawn from a target distribution $\mathbb{P}_T(\x,y)$.
Our goal is to learn a classifier that correctly predicts the labels of $\mathcal{T}$. 


\subsection{Label Alignment with Deep Features}

The proposed label alignment method shares the theoretical motivation of (adversarial learning) methods for domain adaptation: find a common representation that reduces the distance between source and target domain distributions  \cite{ben2007analysis}. In LAD, as in other adversarial adaptation methods, domain distance is reduced using a neural network (the domain discriminator). In LAD the inputs of the network are class probabilities (computed by softmax), while in other adversarial methods, like Domain-Adversarial Neural Networks \cite{ganin2016domain}, the input of the network are (deep) features.


The proposed method tries to align source and target domain at class label level using two models:
 a label classifier $\f(x,\theta_\f)$,
 and a domain discriminator $\g(y,\theta_\g)$.
Both of these functions are parameterized by neural networks, $\theta_\f$ and $\theta_\g$.
For brevity we will omit the parameters $\theta_\f$ and $\theta_\g$.
The label classifier is trained to minimize the following standard supervised loss on the source domain $\mathcal{S}$:
\begin{equation}
  L_S(\f) = \frac{1}{|\mathcal{S}|} \sum_{(\x,y)\in \mathcal{S}} \ell(\f(\x),y),
\end{equation}
where $\ell(p,y)= -\sum_j y_j \log(p_j)$ is the cross-entropy loss, and 
$\x$ denotes the vector of features generated using a pre-trained deep neural network.

The domain discriminator is trained to distinguish the uncertainty of the predictions that $\f$ makes on the source domain from the uncertainty of the predictions on the target domain $\mathcal{T}$.
This is again a standard supervised problem, predicting $d=1$ for the source and $d=0$ for the target, given the output $\f(\x)$ of the label classifier. The loss is
\begin{multline}
  L_D(\f,\g)
     = \frac{1}{|\S|} \sum_{(\x,y) \in \S} \ell(\g(\f(\x)),1)\\
     + \frac{1}{|\Ti|} \sum_{\x \in \Ti} \ell(\g(\f(\x)),0).
\end{multline}

We want the label classifier to `fool' the domain discriminator. This can be achieved by training it to make the predictions on the two domains indistinguishable. That means that we \emph{maximize} $L_D$ with respect to $\f$. The resulting optimization problem is therefore
\begin{equation}\label{optimization}
  \min_{\f}\ \max_{\g} \  L_S(\f) - L_D(\f,\g). \\
\end{equation}

\subsubsection{Class Weighted Loss Function}
\label{sec:weighted}

Since here we adversarially train a domain invariant label classifier on the level of predictions, it is necessary that the label distribution of both domains is the same, otherwise predictions towards certain labels could be based on possible differences between label occurrences in both domains. 

This would have a negative effect and prevent domain alignment.
Formally, let $c_S(y) = |\{(x',y') \in \S \mid y=y'\}|/|\S|$ be the fraction of source instances that have label $y$, and similarly let $c_T(y)$ be the (unknown) fraction of target instances with label $y$ under the true labeling.
Then if $c_S(y) \neq c_T(y)$ and the label classifier $\f$ makes perfect predictions, the domain discriminator is able to distinguish the two domains, based purely on the different conditional probabilities $\mathbb{P}(y|d)$.

To overcome this problem we will switch to weighted loss functions,
\begin{equation}
  L_S(\f) = \frac{1}{|\S|} \sum_{(\x,y)\in \S} w_\S(\x,y) \ell(\f(\x),y),
\end{equation}
\begin{equation}
\begin{split}
  L_D(\f,\g) &= \frac{1}{|\S|} \sum_{(\x,y) \in S} w_S(\x,y) \ell(\g(\f(x)),1) \\
  &+ \frac{1}{|\Ti|} \sum_{\x \in \Ti} w_T(\x) \ell(\g(\f(\x)),0).
\end{split}
\end{equation}
Where the weights for the source domain are
\begin{equation}
  \label{eq:weight_src}
  w_S(\x,y) = \frac{\max_{y'} c_S(y') }{ c_S(y) }.
\end{equation}
For the target domain we do not know the true labels, so instead we use the predicted (pseudo)labels $\tilde{y}(\x) = \argmax_i \f(\x)_i$.
So the weights are
\begin{equation}
  \label{eq:weight_tgt}
  w_T(\x) = \frac{\max_{y'} \tilde{c}_T(y') }{ \tilde{c}_T(\tilde{y}(\x)) },
\end{equation}
where
\begin{equation*}
  \tilde{c}_T(y) = |\{\x' \in T \mid y=\tilde{y}(\x')\}| / |\Ti|.
\end{equation*}

With the weighted loss, the domain discriminator cannot use the difference in conditional probability of the class given the domain, since all classes occur with the same total weight in both domains.

\subsection{Architecture}
The overall LAD architecture used in our experiments is shown in \figref{fig:architecture}. It consists of three parts: feature extractor, label classifier and domain discriminator.

\subsubsection{Feature Extractor}
We use a deep neural network pre-trained on the ImageNet dataset \cite{russakovsky2015imagenet}. The last label prediction layer of a pre-trained network is omitted and features are extracted from the second to last layer, as this is presumably the layer with the lowest maximum mean discrepancy \cite{tzeng2014deep}.   

To generate robust features, we use a form of data augmentation, where different crops and flips of each image are passed through the network, and the features are averaged.

In particular, for each image, its features are calculated as follows.
First, we resize the input image to the input size of the network plus $64$ pixels (for example, for ResNet50, which expects a $224\times224$ input, we resize the image to $288\times288$ pixels).
From this resized image we take $9$ crops spaced of $32$ pixels apart. This is repeated for the horizontally flipped input image, resulting in $18$ different image crops. For each image, crop features are extracted from the pre-trained network. The final features of the input image are the averaged features of its $18$ crops.


\subsubsection{Label Classifier}
We consider a label classifier consisting of two dense (fully connected) layers of size $1024$ with ReLu activation and $0.5$ dropout \cite{srivastava2014dropout}, followed by a dense layer with softmax activation for label predictions. 

\subsubsection{Domain Discriminator}
The considered domain discriminator has the same structure as the label classifier, but without dropout layers.
The domain discriminator is placed after the softmax layer of the label classifier, and behind a gradient reversal layer \cite{ganin2015unsupervised,ganin2016domain} which acts as an identity function on forward passes through the network, and reverses the gradient on backward passes.
This ensures that we can use the gradient of $L_D$ to simultaneously maximize with respect to $f$ and minimize with respect to $g$ in our optimization problem (\ref{optimization}).

\subsection{Training}
All training is done with minibatch Stochastic Gradient Descent (SGD) with Nesterov momentum. Both the label and domain loss is calculated with categorical cross-entropy. For training, we assume that we already extracted the features from a pre-trained deep neural network. The training of LAD is different from that of normal feedforward neural networks due to having two instead of one loss function.
Each training step we draw a minibatch from both domains without replacement, append the domain identifier and, for the source domain, the class labels. With these inputs, training proceeds as follows: first, the source domain batch is used to train the label classifier, then the source and domain batches are concatenated and together are used to train the domain discriminator.
We call one pass through the source domain an epoch.

The weights $w_T$ for the target domain are recomputed once per epoch. In the first epoch we set the weights to $1$.

The complete training approach is displayed in Algorithm~\ref{algo:training}. 

A technical concern of this training procedure is that as the labels for target domain data are unknown, in the proposed method, the weights for each target domain instance are estimated based on the predicted labels, and then updated iteratively epoch by epoch.  However, there is no guarantee that the iterative procedure is able to find an optimal solution for $w_T$. That means the estimation of $w_T$ may become worse and worse.  Nevertheless, under the assumption that features extracted from the pre-trained deep neural network are well transferable, and that source and target domains are related, this phenomenon should  not happen. This is indeed the case in practice, as substantiated  by results of our extensive empirical analysis.

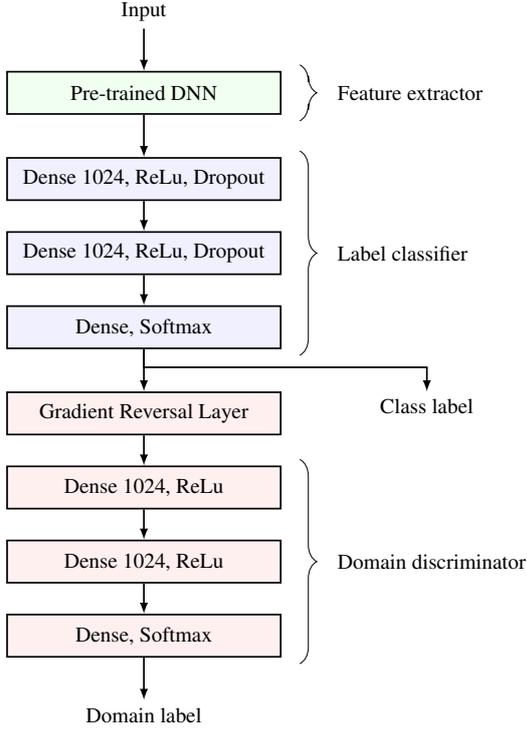
\begin{figure}[!tb]
  \centering
  \tikzset{
    node/.style={minimum height=7mm,thick},
    data/.style={},
    layer/.style={node,draw,minimum width=45mm},
    layer0/.style={layer,fill=green!6!white},
    layer1/.style={layer,fill=blue!6!white},
    layer2/.style={layer,fill=red!6!white},
    layer4/.style={layer,fill=red!6!white},
    far/.style={on chain=going below, node distance=7mm and 15mm},
  }
  \scalebox{0.8}{%
  \begin{tikzpicture}[
      node distance=5mm and 10mm,
      every on chain/.style={join=by -latex, thick},
      every join/.style={thick},
      start chain=net going below
    ]
    \node[on chain, data] (input) {Input};
    \node[on chain, layer0, far] (pre) {Pre-trained DNN};
    \node[on chain, layer1, far] (lc1) {Dense 1024, ReLu, Dropout};
    \node[on chain, layer1] {Dense 1024, ReLu, Dropout};
    \node[on chain, layer1] (lc2) {Dense, Softmax};
    \begin{scope}[start branch=softmax going below right, far]
      \node[on chain, data, every on chain/.style={}] (lbl) {Class label};
    \end{scope}
    \node[on chain, layer4, far] (ld1) {Gradient Reversal Layer};
    \node[on chain, layer2] (ld1) {Dense 1024, ReLu};
    \node[on chain, layer2] {Dense 1024, ReLu};
    \node[on chain, layer2] (ld2) {Dense, Softmax};
    \node[on chain, data, far] {Domain label};
    
    \draw [-latex,thick] (lc2.south) -- +(0,-3mm) -| (lbl);
    \draw [decorate,decoration={brace,amplitude=8pt}] ($(pre.north east) + (3mm,1mm)$) -- ($(pre.south east) + (3mm,-1mm)$) node [black,midway,xshift=5mm,right] {Feature extractor};
    \draw [decorate,decoration={brace,amplitude=8pt}] ($(lc1.north east) + (3mm,1mm)$) -- ($(lc2.south east) + (3mm,-1mm)$)  node [black,midway,xshift=5mm,right] {Label classifier};
    \draw [decorate,decoration={brace,amplitude=8pt}] ($(ld1.north east) + (3mm,1mm)$) -- ($(ld2.south east) + (3mm,-1mm)$)  node [black,midway,xshift=5mm,right] {Domain discriminator};
  \end{tikzpicture}}
  \caption{LAD Architecture}
  \label{fig:architecture}
\end{figure}

\begin{algorithm}[!tb]
   \caption{LAD.}
   \label{algo:training}
   \newcommand{\gen}[1]{G_#1}
\begin{algorithmic}
   \STATE {\bfseries Data:} $\S$ = labeled source data, $\Ti$ = unlabeled target data
   \STATE {\bfseries Result:} $Y$ = predicted labels for target domain
   \STATE $w_S(\x,y) \gets \bigl(\max_{y'} c_S(y')\bigr) / c_S(y)$ for each $(\x,y)\in \S$
   \STATE $w_T(\x) \gets 1$ for each $\x \in \Ti$
   \FOR{$epoch \gets 1,2,\dotsc,n_\text{epochs}$}
   \WHILE{\text{available batches in $\S$}}
   \STATE $S_\text{batch} \gets \text{take batch from } {\S}$
   \STATE $T_\text{batch} \gets \text{take batch from } {\Ti}$
   \STATE Perform a step of SGD on $L_S(\f)$ 
   \STATE Perform a step of SGD on $L_D(\f,\g)$ 
   \ENDWHILE
   \STATE $Y \gets \tilde{y}(\Ti)$ 
   \STATE $w_T(\x) \gets \bigl(\max_{y'} c_T(y')\bigr) / \tilde c_T((\x))$ for each $\x\in \Ti$
   \ENDFOR
\end{algorithmic}
\end{algorithm}

\section{EXPERIMENTS}\label{experiments}

We conduct extensive experiments on $12$ adaptation tasks from two real-life benchmark datasets. Datasets, experimental setup and methods used in our comparative analysis are described in detail below.

\subsection{Datasets}
We consider two benchmark datasets: Office-31 \cite{saenko2010adapting}  and imageCLEF-DA\footnote{\url{http://imageclef.org/2014/adaptation}}.

The Office-31 dataset for visual domain adaptation consists of three domains with images in $31$ categories. The Amazon (A) domain with $2817$ images consists of images taken from Amazon.com product pages. The DSLR (D) and Webcam (W) domains, with respectively $498$ and $795$ images, consist of images taken with either a digital SLR or web camera of the products in different environments. The images in each domain are unbalanced across the $31$ categories, therefore we will use our data balancing method. We report results on all possible domain combinations A$\rightarrow$D, A$\rightarrow$W, D$\rightarrow$A, D$\rightarrow$W, W$\rightarrow$A, and W$\rightarrow$D which is a good combination of difficult and easier domain adaptation tasks.

The imageCLEF-DA dataset is a benchmark dataset for ImageCLEF 2014 domain adaptation challenge and consists of $12$ common categories shared by three public datasets which are seen as different domains: Caltech-256 (C), ImageNet ILSVRC 2012 (I), and Pascal VOC 2012 (P). This dataset is balanced, with $50$ images for each of the $12$ categories for a total of $600$ images per domain, making a good addition to the Office-31 dataset. Since for each transfer task the source is balanced, we omit our own balancing method when using this dataset. We report results on all domain combinations: C$\rightarrow$I, C$\rightarrow$P, I$\rightarrow$C, I$\rightarrow$P, P$\rightarrow$C, P$\rightarrow$I.

\subsection{Experimental Setup}
LAD is implemented on the Tensorflow \cite{tensorflow2015-whitepaper} framework via the Keras \cite{chollet2015keras} interface.  The network and training parameters are kept similar across all pre-trained architectures and domain adaptation tasks of both datasets. Specifically, we use stochastic gradient descent with a learning rate of $0.001$ and Nesterov momentum of $0.9$, a batch size of $32$.  All of these parameter settings are considered default settings. 
In all our experiments we train each model for $n_\text{epochs}=1000$ epochs. For each transfer task we run LAD  $10$ times and report the average label classification accuracy and standard deviation.  

All algorithms are assessed in a fully transductive setup where all unlabeled target instances are used during training for predicting their labels. Labeled instances of the first domain are used as the source and unlabeled instances of the second domain as the target. We evaluate the accuracy on the target domain as the percentage of correctly labeled target instances.

 

In order to assess LAD's transfer capability, we consider a baseline variant, obtained by omitting the domain discriminator from LAD, and trained on the source data (no adaptation). For instance, Baseline(DenseNet201) denotes the baseline variant with the pre-trained DenseNet201 network as the feature extractor. 
Network and training parameters are kept the same as those of LAD across all tasks, besides training for only $100$ epochs which is roughly chosen as optimal before overfitting becomes a problem. 


In all experiments, we did not perform hyperparameter optimization, but just used default settings of Keras. 

\section{RESULTS}

In order to assess comparatively the performance of LAD across different pre-trained architectures, we conduct extensive experiments on the following pre-trained architectures publicly available at Keras: MobileNet \cite{howard2017mobilenets}, VGG16 \cite{simonyan2014very}, VGG19 \cite{simonyan2014very}, DenseNet \cite{huang2017densely}, InceptionV3 \cite{szegedy2016rethinking}, Xception \cite{chollet2016xception}, and InceptionResNetV2 \cite{szegedy2017inception}.

As shown in \Tiblref{table:architectures_office31_weight}, on the Office-31,  LAD(InceptionResNetV2)  outperforms the other variants with an average accuracy of $90.7\%$. Differences between architectures are very clear when looking at their baseline results where the difference between the worst and best architecture is around $10\%$. The InceptionResNetV2 pre-trained features are so good and robust that without LAD they already outperform current state-of-the-art methods for domain adaptation based on the ResNet50 architecture.

On the ResNet50 architecture LAD improves on our baseline (no adaptation) on all tasks. The  improvement is more evident on the harder tasks A$\rightarrow$D, D$\rightarrow$A, A$\rightarrow$W, and W$\rightarrow$A. In particular, on A$\rightarrow$W more than 13\% improvement is achieved (from $76.5$ with no adaptation to $89.9$ with adaptation). 

The increase in target accuracy is larger when using less powerful architectures. For example, with MobileNet, on the harder adaptation tasks D$\rightarrow$A and W$\rightarrow$A, about $15\%$ increase in target accuracy is achieved (from $57.2$ with no adaptation to $72.1$ with adaptation  for D$\rightarrow$A, and from $56.5$ with no adaptation to $71.3$ with adaptation for W$\rightarrow$A).

As shown in \Tiblref{table:architectures_imageclef_weight}, on the ImageClef-DA adaptation tasks, the best average accuracy is obtained by LAD with the Xception  architecture, with an average accuracy of $89.68\%$. Notably, on the C$\rightarrow$I adaptation task,  using InceptionResNetV2 LAD  gains about $11\%$ target accuracy over the Baseline (from $80.3$ with no adaptation to $91.5$ with adaptation).

ImageCLEF-DA results of LAD based on ResNet50 show that the best improvement over the Baseline (no adaptation) is obtained on harder tasks. For instance, on the C$\rightarrow$I task (from $80.9$ with no adaptation to $88.5$ with adaptation). 

LAD consistently performs well on features from  pre-trained deep neural networks with different architectures. 

Overall, results indicate that more recent pre-trained models achieve very good performance and that LAD consistently improves on the baselines. 
These results provide further experimental evidence  that deep networks learn feature representations which reduce domain discrepancy, but do not fully eliminate it, even for architectures achieving excellent performance, like InceptionResNetV2.


\begin{table*}[!htb]
 \small{}
 \centering
 \caption{Baseline and LAD average accuracy (with standard deviations) over $10$ runs on the Office-31 dataset for different network architectures.}
 \label{table:architectures_office31_weight}
 \begin{tabular}{|p{4cm}|ccccccc|}
 \hline
 Method                           & A $\rightarrow$ D            & A $\rightarrow$ W             & D $\rightarrow$ A             & D $\rightarrow$ W             & W $\rightarrow$ A             & W $\rightarrow$ D             & avg               \\ \hline
 Baseline(MobileNet)              & 74.5$\pm$1.5               & 73.5$\pm$0.6                & 57.2$\pm$0.6                & 97.8$\pm$0.2                & 56.5$\pm$0.6                & 99.4$\pm$0.2                & 76.5\%           \\
 LAD(MobileNet)                 & 82.2$\pm$1.8               & 89.3$\pm$2.5                & 72.1$\pm$0.5                & \textbf{98.9$\pm$0.1}       & 71.3$\pm$3.2                & 99.8$\pm$0.1                & 85.6\%           \\ \hline
 Baseline(VGG16)                  & 76.5$\pm$1.1               & 73.7$\pm$1.2                & 61.9$\pm$0.6                & 96.6$\pm$0.3                & 60.4$\pm$0.6                & 99.7$\pm$0.1                & 78.1\%           \\
 LAD(VGG16)                     & 85.3$\pm$2.0               & 87.9$\pm$1.5                & 69.9$\pm$0.8                & 97.3$\pm$0.2                & 70.1$\pm$0.6                & 99.7$\pm$0.1                & 85.0\%           \\ \hline
 Baseline(VGG19)                  & 76.1$\pm$0.8               & 72.9$\pm$1.1                & 63.4$\pm$0.6                & 97.4$\pm$0.4                & 62.9$\pm$1.0                & 99.8$\pm$0.1                & 78.8\%           \\
 LAD(VGG19)                     & 83.9$\pm$1.8               & 87.7$\pm$0.7                & 71.0$\pm$0.8                & 98.2$\pm$0.3                & 71.5$\pm$0.8                & 99.9$\pm$0.1                & 85.4\%           \\ \hline
 Baseline(ResNet50)               & 81.0$\pm$0.6               & 76.5$\pm$0.9                & 64.8$\pm$0.8                & 97.5$\pm$0.2                & 63.6$\pm$1.0                & 99.7$\pm$0.2                & 80.5\%           \\
 LAD(ResNet50)                  & 90.6$\pm$1.2               & 90.0$\pm$0.7                & 74.0$\pm$0.6                & 98.0$\pm$0.1                & 75.3$\pm$1.4                & 99.8$\pm$0.2                & 87.9\%           \\ \hline
 Baseline(DenseNet201)            & 85.3$\pm$0.8               & 82.3$\pm$1.2                & 68.5$\pm$0.6                & 98.0$\pm$0.2                & 67.7$\pm$0.5                & 99.9$\pm$0.1                & 83.6\%           \\
 LAD(DenseNet201)               & 93.1$\pm$0.8               & 94.7$\pm$0.9                & 77.2$\pm$0.8                & 98.6$\pm$0.1                & 77.7$\pm$0.7                & 99.9$\pm$0.1                & 90.2\%           \\ \hline
 Baseline(InceptionV3)            & 85.9$\pm$0.8               & 82.4$\pm$0.7                & 72.8$\pm$0.4                & 97.5$\pm$0.4                & 72.8$\pm$0.3                & 99.0$\pm$0.3                & 85.1\%           \\
 LAD(InceptionV3)               & 91.2$\pm$0.7               & 88.6$\pm$0.5                & 76.9$\pm$0.5                & 98.3$\pm$0.2                & 76.9$\pm$0.8                & 99.3$\pm$0.2                & 88.5\%           \\ \hline
 Baseline(Xception)               & 85.2$\pm$0.7               & 83.9$\pm$0.7                & 72.1$\pm$0.4                & 97.0$\pm$0.2                & 71.9$\pm$0.5                & 99.7$\pm$0.1                & 85.0\%           \\
 LAD(Xception)                  & 91.0$\pm$1.5               & 92.9$\pm$0.5                & 78.6$\pm$0.3                & 98.1$\pm$0.1                & 78.1$\pm$0.8                & \textbf{100.0$\pm$0.1}       & 89.8\%           \\ \hline
 Baseline(InceptionResNetV2)      & 90.2$\pm$0.7               & 89.3$\pm$0.6                & 74.9$\pm$0.5                & 97.3$\pm$0.2                & 75.5$\pm$0.3                & 99.6$\pm$0.2                & 87.8\%           \\
 LAD(InceptionResNetV2)         & \textbf{93.7$\pm$0.8}      & \textbf{95.3$\pm$0.3}       & \textbf{78.8$\pm$0.5}       & 98.3$\pm$0.1                & \textbf{78.5$\pm$0.5}       & 99.6$\pm$0.1                & \textbf{90.7\%}  \\ \hline
 \end{tabular}
 \vspace*{3mm}
\end{table*}

\begin{table*}[!htb]
 \centering
 \caption{Baseline and LAD average accuracy (with standard deviations) over $10$ runs on the ImageCLEF-DA dataset for different network architectures.}
 \label{table:architectures_imageclef_weight}
 \small{}
 \begin{tabular}{|p{4cm}|ccccccc|}
 \hline
 Method                           & C $\rightarrow$ I            & C $\rightarrow$ P             & I $\rightarrow$ C             & I $\rightarrow$ P             & P $\rightarrow$ C             & P $\rightarrow$ I             & avg               \\ \hline
 Baseline(MobileNet)              & 77.9$\pm$0.3               & 65.2$\pm$0.8                & 89.8$\pm$0.7                & 74.6$\pm$0.4                & 91.2$\pm$0.8                & 84.9$\pm$0.8                & 80.6\%           \\
 LAD(MobileNet)                 & 87.9$\pm$0.7               & 73.9$\pm$0.7                & 94.6$\pm$0.4                & 75.2$\pm$0.5                & 94.0$\pm$0.3                & 88.3$\pm$0.7                & 85.6\%           \\ \hline
 Baseline(VGG16)                  & 83.2$\pm$0.7               & 70.7$\pm$0.5                & 91.9$\pm$0.5                & 76.5$\pm$0.5                & 91.5$\pm$0.6                & 86.0$\pm$0.8                & 83.3\%           \\
 LAD(VGG16)                     & 89.6$\pm$0.5               & 76.7$\pm$0.8                & 94.3$\pm$0.3                & 76.2$\pm$0.8                & 94.4$\pm$0.4                & 88.8$\pm$0.9                & 86.7\%           \\ \hline
 Baseline(VGG19)                  & 84.7$\pm$0.7               & 70.9$\pm$0.4                & 92.0$\pm$0.3                & 76.6$\pm$0.4                & 91.6$\pm$0.5                & 85.8$\pm$0.7                & 83.6\%           \\
 LAD(VGG19)                     & 89.0$\pm$0.7               & 74.5$\pm$0.5                & 94.8$\pm$0.3                & 77.3$\pm$0.6                & 94.3$\pm$0.3                & 90.2$\pm$1.0                & 86.7\%           \\ \hline
 Baseline(ResNet50)               & 80.9$\pm$1.3               & 68.0$\pm$1.0                & 92.2$\pm$0.5                & 76.1$\pm$0.4                & 91.8$\pm$0.5                & 88.4$\pm$0.8                & 82.9\%           \\
 LAD(ResNet50)                  & 88.5$\pm$1.0               & 74.0$\pm$1.0                & 95.2$\pm$0.4                & 76.8$\pm$0.7                & 94.1$\pm$0.2                & 90.6$\pm$0.6                & 86.5\%           \\ \hline
 Baseline(DenseNet201)            & 87.7$\pm$0.7               & 71.6$\pm$0.6                & 93.6$\pm$0.4                & 78.3$\pm$0.4                & 94.3$\pm$0.5                & 90.8$\pm$0.8                & 86.1\%           \\
 LAD(DenseNet201)               & 93.0$\pm$0.4               & \textbf{78.3$\pm$1.0}       & \textbf{97.5$\pm$0.3}       & 79.1$\pm$0.3                & 95.7$\pm$0.4                & 93.2$\pm$0.4                & 89.5\%           \\ \hline
 Baseline(InceptionV3)            & 83.1$\pm$1.2               & 66.1$\pm$0.8                & 94.3$\pm$0.5                & 77.8$\pm$0.5                & 93.9$\pm$0.4                & 90.8$\pm$0.9                & 84.3\%           \\
 LAD(InceptionV3)               & 92.8$\pm$0.3               & 75.9$\pm$0.9                & 95.9$\pm$0.3                & 78.3$\pm$0.5                & 95.8$\pm$0.3                & \textbf{94.2$\pm$0.5}       & 88.8\%           \\ \hline
 Baseline(Xception)               & 85.2$\pm$0.8               & 69.9$\pm$0.5                & 94.7$\pm$0.5                & 79.3$\pm$0.5                & 92.8$\pm$1.1                & 90.8$\pm$0.6                & 85.5\%           \\
 LAD(Xception)                  & \textbf{94.2$\pm$0.4}      & 77.7$\pm$1.1                & 96.8$\pm$0.4                & 80.1$\pm$0.5                & \textbf{96.6$\pm$0.3}       & 92.6$\pm$0.6                & \textbf{89.7\%}  \\ \hline
 Baseline(InceptionResNetV2)      & 80.3$\pm$0.9               & 67.8$\pm$0.9                & 90.3$\pm$1.9                & 79.3$\pm$0.5                & 88.4$\pm$0.9                & 89.7$\pm$0.8                & 82.6\%           \\
 LAD(InceptionResNetV2)         & 91.5$\pm$0.7               & 75.9$\pm$0.9                & 97.2$\pm$0.3                & \textbf{80.6$\pm$0.5}       & 95.0$\pm$0.3                & 92.3$\pm$1.2                & 88.7\%           \\ \hline
 \end{tabular}
 \vspace*{3mm}
\end{table*}

\section{COMPARISON WITH END-TO-END DEEP LEARNING METHODS}

To assess how results of LAD compare with the state-of-the-art, we report published results of the following end-to-end deep learning methods for domain adaptation that fine-tune a ResNet50 model pre-trained on ImageNet: 
Deep Domain Confusion (DDC) \cite{tzeng2014deep},
Deep Adaptation Network (DAN) \cite{long2015learning},
Residual Transfer Network (RTN) \cite{long2016unsupervised},
Adversarial Discriminative Domain Adaptation (ADDA) \cite{Tzeng_2017_CVPR},
Reverse Gradient (RevGrad)  \cite{ganin2015unsupervised}. 

Although all experiments were conducted under the same transductive setup, results should be interpreted with care. There are various differences between the considered algorithms. For instance, end-to-end training of a pre-trained deep architecture versus using the pre-trained architecture to extract features, or hyper-parameters tuning vs using default settings.  

Overall, results indicate state of the art performance of LAD, comparable or better than that of end-to-end deep adaptation methods.

\begin{table*}[!htb]
\caption{Average accuracy (with standard deviations) on adaptation tasks from the Office-31 dataset. All methods considered use  a ResNet50 model.}
\small{}
\centering
\begin{tabular}{|l|ccccccc|}
\hline
Method                           & A $\rightarrow$ D            & A $\rightarrow$ W            & D $\rightarrow$ A             & D $\rightarrow$ W             & W $\rightarrow$ A             & W $\rightarrow$ D             & avg               \\
\hline
DDC \cite{tzeng2014deep} &77.5$\pm$0.3 &75.8$\pm$0.2 &67.4$\pm$0.4&95.0$\pm$0.2 & 64.0$\pm$0.5 &98.2$\pm$0.1 &79.7\% \\
DAN \cite{long2015learning} &78.4$\pm$0.2 &83.8$\pm$0.4 &66.7$\pm$0.&96.8$\pm$0.2 & 62.7$\pm$0.2 &99.5$\pm$0.1 &81.3\% \\
RTN \cite{long2016unsupervised} &71.0$\pm$0.2 &73.3$\pm$0.2 & 50.5$\pm$0.3 & 96.8$\pm$0.2 &51.0$\pm$0.1 &99.6$\pm$0.1 &73.7\% \\
RevGrad \cite{ganin2015unsupervised} & 72.3$\pm$0.3 &73.0$\pm$0.5 & 52.4$\pm$0.4 &96.4$\pm$0.3 & 50.4$\pm$0.5 & 99.2$\pm$0.3 &74.1\% \\
ADDA \cite{Tzeng_2017_CVPR} & 77.8$\pm$0.3 &86.2$\pm$0.5 &69.5$\pm$0.4& 96.2$\pm$0.3 &68.9$\pm$0.5 &98.4$\pm$0.3 &82.9\%\\
LAD    & \textbf{90.6$\pm$1.2}  & \textbf{89.9$\pm$0.7} & \textbf{74.0$\pm$0.6}   & \textbf{98.0$\pm$0.1} &\textbf{75.3$\pm$1.4}  & \textbf{99.8$\pm$0.2} &  \textbf{87.9\%} \\
\hline
\end{tabular}
 \vspace*{3mm}
\label{tbl:resnet50office31}
\end{table*}





\begin{table*}[!htb]
\centering
\caption{Average accuracy (with standard deviations) for various methods on the ImageCLEF-DA dataset, obtained with the ResNet50 architecture.}
\small{}
\begin{tabular}{|l|ccccccc|}
\hline
Method                           & I $\rightarrow$ P            & P $\rightarrow$ I            & I $\rightarrow$ C             & C $\rightarrow$ I             & C $\rightarrow$ P            & P $\rightarrow$ C             & avg               \\ \hline
DAN \cite{long2015learning} &75.0$\pm$0.4 &86.2$\pm$0.2 &93.3$\pm$0.2 &84.1$\pm$0.4& 69.8$\pm$0.4 &91.3$\pm$0.4& 83.3\%\\
RTN \cite{long2016deep}  &75.6$\pm$0.3 &86.8$\pm$0.1 &95.3$\pm$0.1 &86.9$\pm$0.3& 72.7$\pm$0.3 &92.2$\pm$0.4 &84.9\%\\
RevGrad \cite{ganin2015unsupervised} &75.0$\pm$0.6 &86.0$\pm$0.3 &\textbf{96.2$\pm$0.4}& 87.0$\pm$0.5 &\textbf{74.3$\pm$0.5} &91.5$\pm$0.6 &85.0\%\\

LAD                     &   \textbf{76.8$\pm$0.7  }           &             \textbf{90.6$\pm$0.6}      &       95.2$\pm$0.3           &        \textbf{ 88.5$\pm$1.0}        &    74.0$\pm$1.0    &      \textbf{94.1$\pm$0.2}     & \textbf{86.5\%}  \\ \hline
\end{tabular}
\label{table:resnet50_imageclef}
\end{table*}

\section{DISCUSSION}\label{discussion}

\subsection{Effectiveness with Shallower Pre-Trained Deep Models}
LAD depends on the quality of pseudo labels for computing weights of target instances and for the model construction. A natural concern is: What if target classification accuracy is too low? Will the alignment of classifier predictions still be effective? To investigate this issue, we consider the shallower network AlexNet as feature extractor for the Office-31 dataset. Since this model is not available in Keras, we used deep features from the 7th layer provided by \cite{tommasi2014testbed}.  \Tiblref{table:alexnet_office31} shows results. When using the less deep AlexNet architecture LAD still improves on our baseline (no adaptation) on all tasks. Also in this case, adaptation proves to be effective on harder tasks. For instance on W$\rightarrow$A our baseline obtains $46.1$ accuracy, while with adaptation $54.8$ accuracy is achieved. 

\begin{table*}[tb]
\centering
\caption{Average accuracy (with standard deviations) on adaptation tasks from the Office-31 dataset. LAD uses features extracted from the 7th layer of the pre-trained AlexNet  model. }
\small{}
\begin{tabular}{|l|ccccccc|}
\hline
Method                           & A $\rightarrow$ D            & A $\rightarrow$ W            & D $\rightarrow$ A             & D $\rightarrow$ W             & W $\rightarrow$ A             & W $\rightarrow$ D             & avg               \\
\hline
Baseline(DeCAF-fc7)              & 63.63$\pm$1.07               & 57.26$\pm$1.17                & 47.53$\pm$0.75                & 94.30$\pm$0.66                & 46.15$\pm$0.61                & 98.07$\pm$0.42                & 67.82\%           \\
LAD(DeCAF-fc7)                 & \textbf{70.78$\pm$1.25}               & \textbf{65.77$\pm$0.56}                & \textbf{53.47$\pm$0.96}                & \textbf{96.78$\pm$0.39}                & \textbf{54.82$\pm$1.18}                & \textbf{98.94$\pm$0.32}                & \textbf{73.43}\%           \\
\hline
\end{tabular}
\label{table:alexnet_office31}
\end{table*}

\subsection{Robustness to the Choice of the Number of Epochs}



Looking at the learning curves in \figref{fig:robustness}, we see that the target domain classification loss reaches a minimum after 50 to 150 epochs, after which it starts to increase. However, the accuracy continues to increase, and there is no sign of overfitting.
Ganin \& Lempitsky \cite{ganin2015unsupervised} also report this finding for their method, but it seems this phenomenon is even more pronounced when aligning domains on the level of predictions instead of features.
Indeed, aligning domains on predictions needs to entail the same level of certainty of predictions for both source and target domains, which leads to an overestimation of the target domain prediction certainty, to match the certainty on the source domain. This overestimation in time results in an increased loss while stabilizing accuracy: a higher certainty of target predictions makes it harder to switch predictions to another class label. 

Furthermore, while the certainty on the source domain leads to overconfidence of the label classifier on the target domain, the uncertainty about the target domain labels has a regularizing effect on the source domain. The label classifier cannot become overconfident on the source domain, because then the source domain predictions would not look like the initially uncertain target domain predictions.

The stability of the target domain, together with the regularizing effect of the label uncertainty on the source domain makes LAD robust to the choice of the number of epochs. The algorithm therefore does not require early stopping.

\begingroup
\centering
\begin{figure}[htbp]
\centering
\subfigure[Office-31 accuracy.]{\begin{tikzpicture}[scale=0.75]
      \begin{axis}[xmin=0, xmax=1000, ylabel={accuracy (\%)}, xlabel={epochs}, width=6cm]
      \addplot[blue] 
      coordinates {(10,80.26)(20,81.58)(30,82.42)(40,83.18)(50,83.58)(60,84.24)(70,84.54)(80,85.00)(90,85.29)(100,85.59)(120,85.99)(140,86.41)(160,86.56)(180,86.80)(200,86.99)(250,87.16)(300,87.40)(350,87.53)(400,87.55)(450,87.54)(500,87.58)(600,87.64)(700,87.67)(800,87.69)(900,87.76)(1000,87.75)};
      \end{axis}
    \end{tikzpicture}}\label{fig:1a}
\subfigure[Office-31 loss.] {\begin{tikzpicture}[scale=0.75]
      \begin{axis}[xmin=0, xmax=1000, ylabel={class loss}, xlabel={epochs}, width=6cm]
      \addplot[red] 
      coordinates {(10,0.77)(20,0.69)(30,0.67)(40,0.65)(50,0.63)(60,0.62)(70,0.61)(80,0.60)(90,0.60)(100,0.59)(120,0.59)(140,0.58)(160,0.59)(180,0.59)(200,0.59)(250,0.60)(300,0.61)(350,0.62)(400,0.64)(450,0.66)(500,0.67)(600,0.71)(700,0.74)(800,0.77)(900,0.78)(1000,0.80)};
      \end{axis}
    \end{tikzpicture}}\label{fig:1b}
\subfigure[ImageCLEF-DA accuracy.]{\begin{tikzpicture}[scale=0.75]
      \begin{axis}[xmin=0, xmax=1000, ylabel={accuracy (\%)}, xlabel={epochs}, width=6cm]
      \addplot[blue] 
      coordinates {(10,81.96)(20,82.64)(30,83.27)(40,83.97)(50,84.35)(60,84.76)(70,84.99)(80,85.31)(90,85.41)(100,85.67)(120,86.05)(140,86.34)(160,86.56)(180,86.76)(200,86.79)(250,86.93)(300,86.92)(350,87.04)(400,87.06)(450,87.01)(500,87.05)(600,87.05)(700,87.05)(800,87.11)(900,87.14)(1000,87.15)};
      \end{axis}
    \end{tikzpicture}}\label{fig:1c}
\subfigure[ImageCLEF-DA loss.]{ \begin{tikzpicture}[scale=0.75]
      \begin{axis}[xmin=0, xmax=1000, ylabel={class loss}, xlabel={epochs}, width=6cm]
      \addplot[red] 
      coordinates {(10,0.68)(20,0.67)(30,0.67)(40,0.67)(50,0.67)(60,0.68)(70,0.68)(80,0.69)(90,0.69)(100,0.70)(120,0.71)(140,0.73)(160,0.75)(180,0.76)(200,0.77)(250,0.81)(300,0.84)(350,0.86)(400,0.88)(450,0.90)(500,0.91)(600,0.93)(700,0.96)(800,0.98)(900,1.01)(1000,1.04)};
      \end{axis}
    \end{tikzpicture}}\label{fig:1d}
\caption{Target domain classification accuracy and classification loss when training for up to $1000$ epochs.
  Made with the ResNet50 architecture.}\label{fig:robustness}
\end{figure}
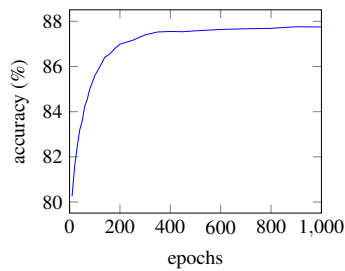
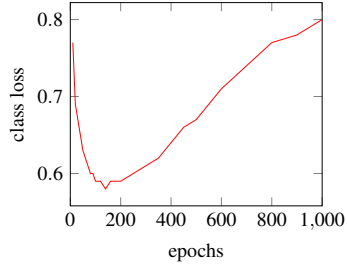
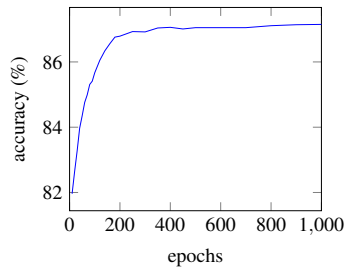
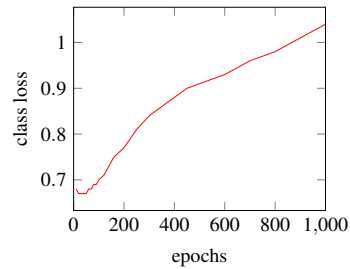
\endgroup

\begingroup
\centering
\begin{figure}[htbp]
\centering
\subfigure[Baseline source.]{\includegraphics[width=0.23\textwidth]{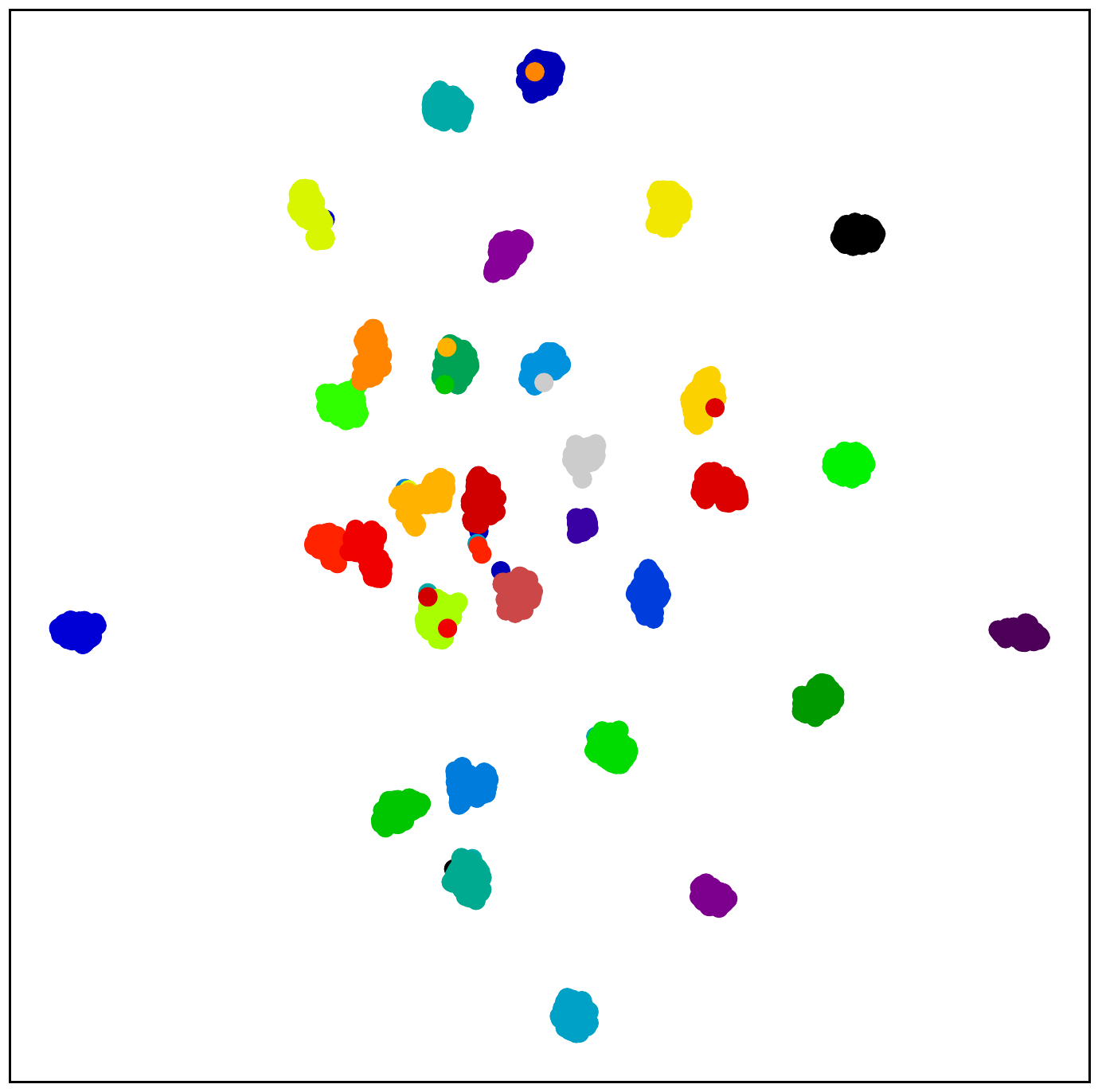}}\label{fig:2a}
\subfigure[Baseline target.] {\includegraphics[width=0.23\textwidth]{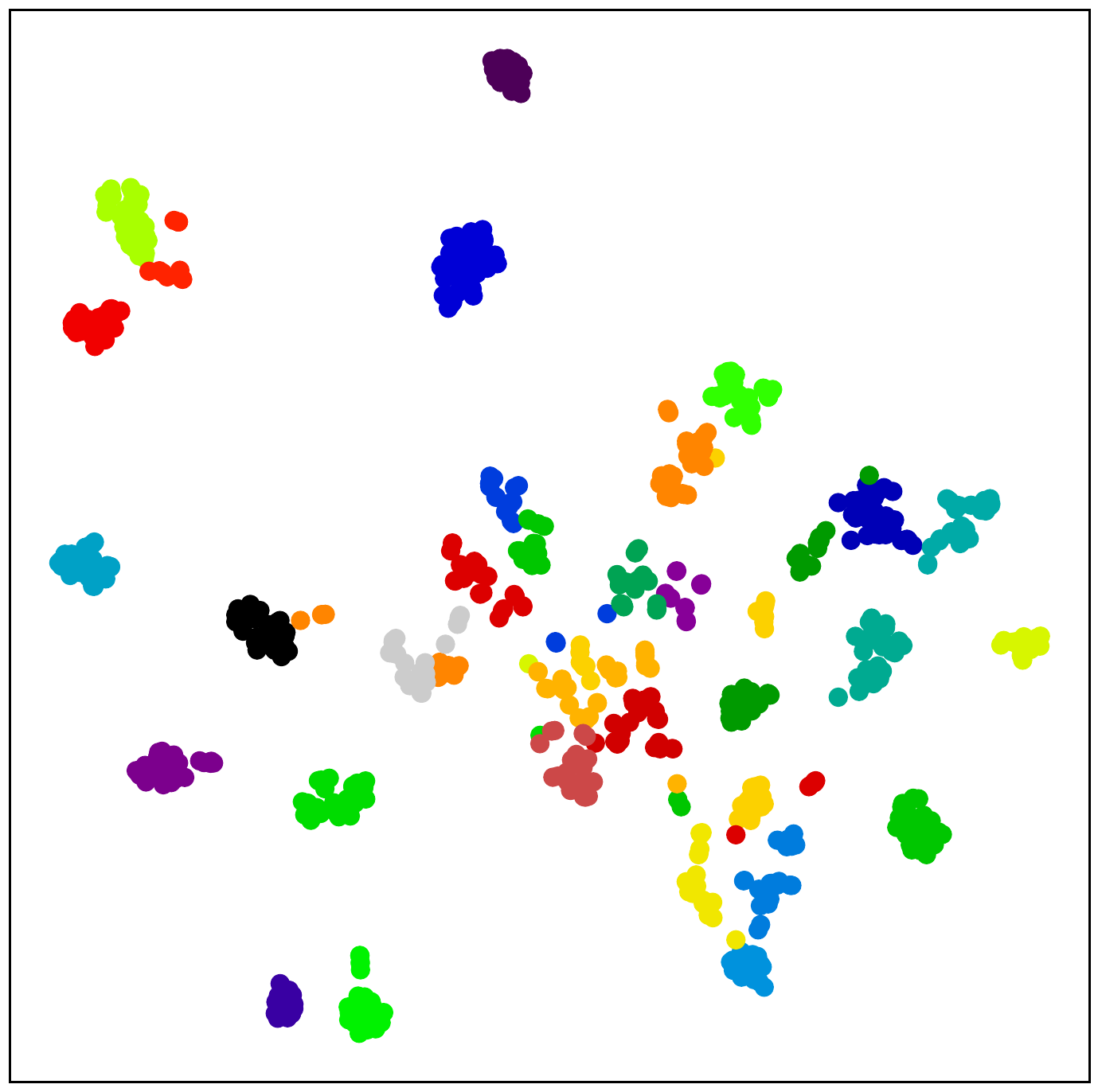}}\label{fig:2b}
\subfigure[LAD source.]{\includegraphics[width=0.23\textwidth]{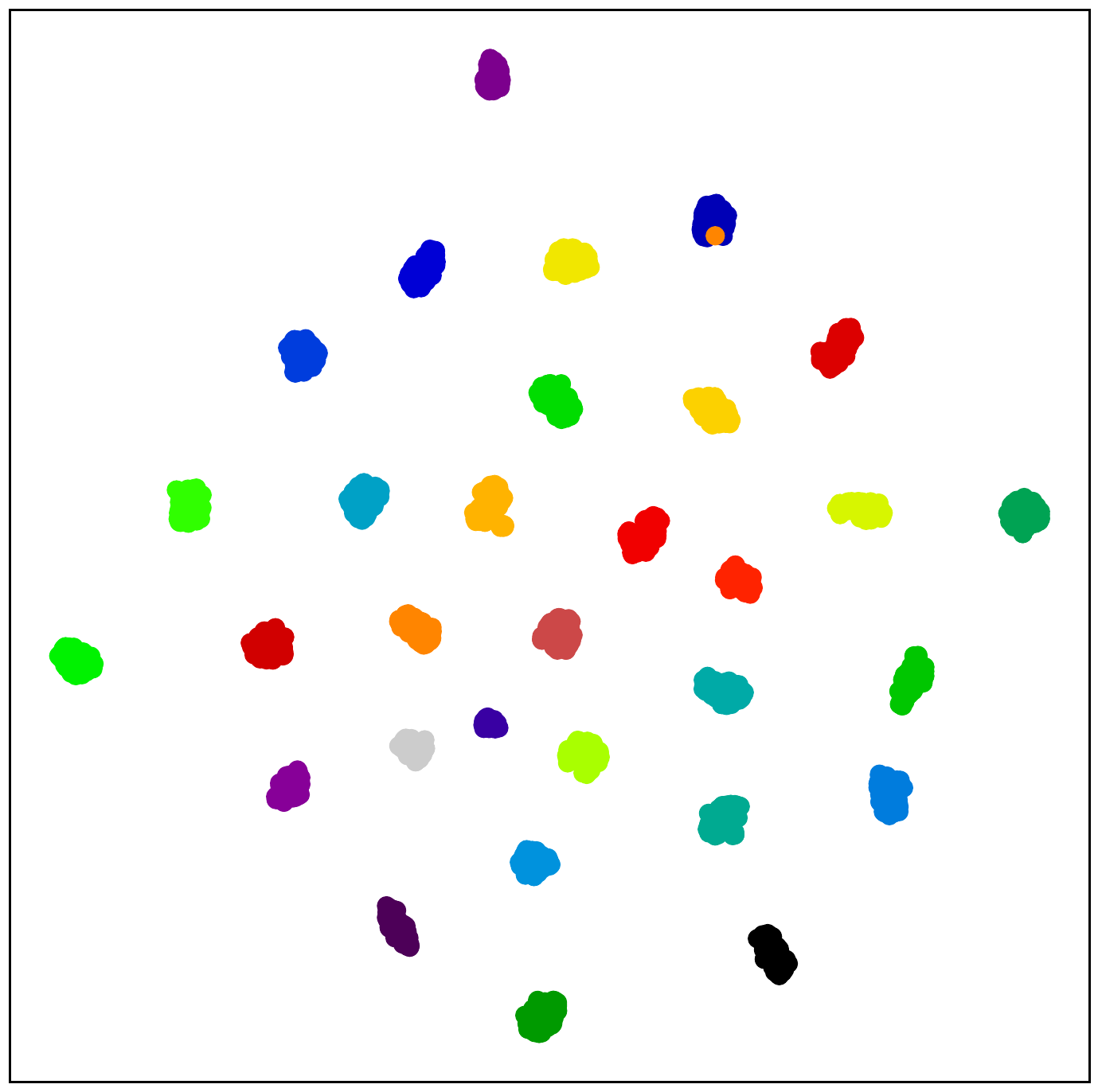}}\label{fig:2c}
\subfigure[LAD target.]{\includegraphics[width=0.23\textwidth]{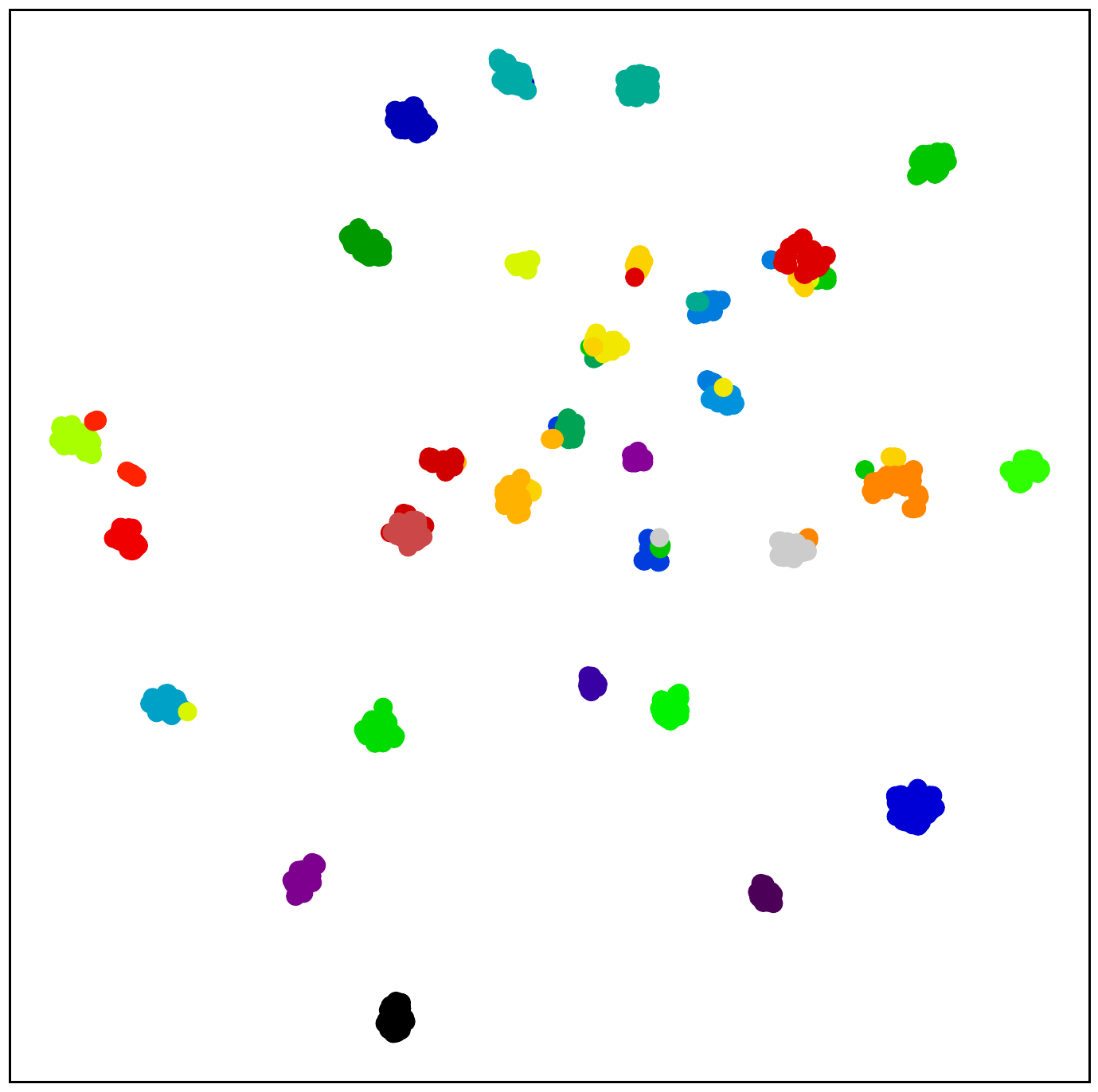}}\label{fig:2d}
\caption{t-SNE feature visualization of Baseline and LAD features on the A$\to$W task from the Office-31 dataset. ResNet50 is the used pre-trained architecture. Features visualized from the second dense layer of our architecture shown in \figref{fig:architecture}.}\label{fig:tsne}
\end{figure}
\endgroup

%

\subsection{Class Weights Importance with Unbalanced Class Distributions}


%
%


We have also investigated the importance of the class weights introduced in our loss function  (see \secref{sec:weighted}), by training the model without using weights.

On the Office-31 dataset, without class weights LAD with ResNet50 features achieves an average accuracy of 80.3\%, compared to 87.9\% when the  loss with class weights is used.
The Office-31 dataset has unbalanced class distributions. In this case the  loss with weights prevents the use of this information, and LAD obtains better performance.

On the other hand, on ImageCLEF-DA, not using class weights gives an average accuracy of 87.8\%, compared to 86.5\% with weights.
This happens because this dataset is fully class balanced. In that case, performance does not drop when no weights are used in the loss, because class distributions are already fully class balanced.

In general, we can make no assumptions about the target domain being balanced. In that case, we should assume that the data is not class balanced, and use the weighted loss functions, as is done in LAD.


\subsection{Running Time}
LAD does not perform fine-tuning of large pre-trained architecture weights and therefore is relatively fast to train. On average over all different transfer tasks a single epoch as described in algorithm \ref{algo:training} takes $0.4$ seconds for Office-31 and $0.2$ seconds for the imageCLEF-DA dataset when trained on a single Nvidia GeForce GTX 1070.

\subsection{Visualization of Deep Features}
To get more insight into the feature representation learned with LAD, we compare t-SNE \cite{maaten2008visualizing} feature visualizations of LAD features with those of Baseline on the ResNet50 architecture. For better comparability, we visualize features on the difficult A$\rightarrow$W adaptation task. Visualized features are from the second dense layer (see \figref{fig:architecture}). \figref{fig:tsne} indicates that LAD features are better and more domain invariant than those of the baseline, since the $31$ classes of the Office-31 dataset are better distinguishable and the features from both domains are better mapped on each other.

\section{CONCLUSION}\label{conclusion}
In this paper we introduced domain alignment at prediction uncertainty level, to be used with features extracted from pre-trained deep neural networks. We demonstrated effectiveness, efficiency, and robustness through extensive experiments with diverse pre-trained architectures and unsupervised domain adaptation tasks for image classification.  

In our experimental analysis, we did not perform hyperparameter optimization, but just used default settings of Keras. It is interesting to investigate whether LAD performance could be further improved by applying procedures for tuning hyperparameters in a transfer learning setting, like \cite{zhong2010cross}. 

We have shown that training with our tailored loss function favors robustness, because the domain discriminator punishes overconfidence on the source domain, the latter being a sign of overfitting. It will be interesting to investigate whether a similar technique can also be used to prevent overfitting in other settings, such as supervised learning. 

A limitation and intrinsic characteristic of LAD  is that it does not directly align source and target features, it does alignment only through the uncertainty of predictions.  This is a direct consequence of the domain adaptation scenario investigated here. As a consequence, LAD is sensitive to the choice of the features. Although the results of our experiments showed that in practice LAD works well across features from various pre-trained deep neural networks, its underlying assumption is the existence (and availability) of transferable (deep) features. On the other hand, domain alignment at the feature level as performed by previous domain adaptation methods, notably RevGrad, does not rely on this assumption and is therefore of more general applicability.

Nevertheless,  our method for prediction uncertainty alignment can be applied to any feature representation that is good for source and target, so it is not limited to pre-trained deep neural networks as  feature extractors.  It will be interesting in future work to explore the utility of the method when used on the top of domain adaptation methods based on feature transformation, like \cite{Fernando2013SA,sun2016return}.

\bibliographystyle{apalike}
{\small
\bibliography{main,da}}
%
%
%

\vfill
\end{document}